\documentclass[conference]{IEEEtran}
\usepackage{amsmath,graphicx,multirow,cite,setspace,array,graphics,balance}
\usepackage{hhline}
\usepackage{caption}
\usepackage{subcaption}
\usepackage[export]{adjustbox}
\usepackage{blindtext,graphicx}
\usepackage[absolute]{textpos}

\makeatletter

\newcommand{\Rmnum}[1]{\expandafter\@slowromancap\romannumeral #1@}
\makeatother

\ifCLASSINFOpdf
\else
\fi
\begin{document}
%
\begin{textblock*}{150mm}(5mm,5mm)
\noindent\normalsize A slightly different version of this paper is accepted at MMSP 2018
\end{textblock*}

\title{A Cloud Detection Algorithm for Remote Sensing Images Using Fully Convolutional Neural Networks}

\author{\IEEEauthorblockN{Sorour Mohajerani, Thomas A. Krammer, Parvaneh Saeedi}
\IEEEauthorblockA{School of Engineering Science\\Simon Fraser University, Burnaby, BC, Canada\\
Email: \{smohajer,tkrammer,psaeedi\}@sfu.ca}
}


%


\maketitle

\begin{abstract}
This paper presents a deep-learning based framework for addressing the problem of accurate cloud detection in remote sensing images. This framework benefits from a Fully Convolutional Neural Network (FCN), which is capable of pixel-level labeling of cloud regions in a Landsat 8 image. Also, a gradient-based identification approach is proposed to identify and exclude regions of snow/ice in the ground truths of the training set. We show that using the hybrid of the two methods (threshold-based and deep-learning) improves the performance of the cloud identification process without the need to manually correct automatically generated ground truths. In average the \textit{Jaccard} index and recall measure are improved by 4.36\% and 3.62\%, respectively.

\end{abstract}

\begin{IEEEkeywords}
Cloud detection, remote sensing, Landsat 8, image segmentation, deep-learning, CNN, FCN, U-Net. 
\end{IEEEkeywords}

%
\IEEEpeerreviewmaketitle

\section{Introduction}
Creating an accurate measure of cloud cover is a crucial step in the collection of satellite imagery. The presence of cloud and its coverage level in an image could affect the integrity and the value of that image in most remote sensing applications that rely on optical satellite imagery. Moreover, transmission and storage of images with high cloud coverage seem to be unnecessary and perhaps even wasteful.  Therefore, accurate identification of cloud regions in satellite images is an active subject of research.  Since clouds share similar reflection characteristics with some other ground objects/surfaces such as snow, ice, and white man-made objects, identification of the cloud and its separation from non-cloud regions is a challenging task. The existence of additional data such as multi-spectral bands could assist a more accurate cloud identification process by utilizing temperature and water content information that are provided through additional bands. The difficulty in automation of cloud segmentation becomes more significant when access to spectral bands is limited to Red, Green, Blue, and Near-infrared (Nir) only. Such limitation exists in the data of many satellites such as HJ-1 and GF-2, as they do not provide more spectral band data.

In recent years, many cloud detection algorithms have been developed. These methods can be divided into three main categories: threshold-based approaches \cite{fmask,acca}, handcrafted approaches \cite{hot,bag}, and deep-learning based \cite{multilevel}. 

\begin{figure}[t]
\centering

\begin{minipage}{0.2\textwidth}
\centering
\centerline{\includegraphics[height=43mm, width=43mm]{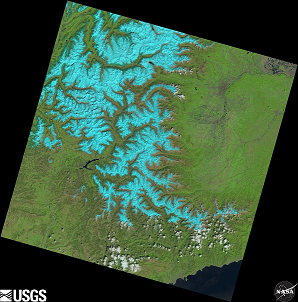}}\footnotesize{(a)}
\end{minipage}
\vspace{1mm}
\hspace{5mm}
\begin{minipage}{0.2\textwidth}
  \centering
  \centerline{\includegraphics[height=43mm, width=43mm]{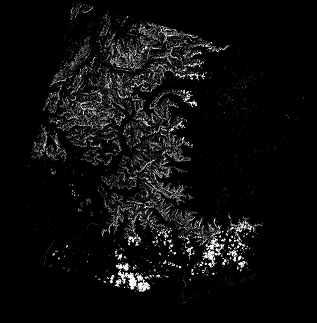}}\footnotesize{(b)}
\end{minipage}

\begin{minipage}{0.2\textwidth}
  \centering
  \centerline{\includegraphics[height=42mm, width=42mm]{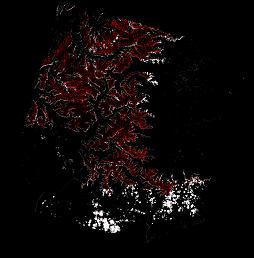}}
\footnotesize{(c)}
  \end{minipage}
\vspace{1mm}
\hspace{5mm}
\begin{minipage}{0.2\textwidth}
  \centering
  \centerline{\includegraphics[height=42mm, width=42mm]{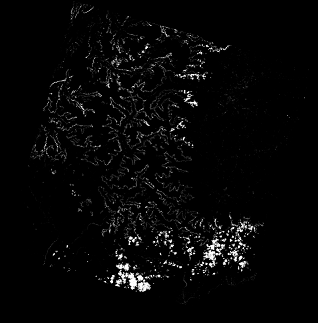}}
\footnotesize{(d)}
  \end{minipage}
\vspace{-1mm}
\hspace{-2.5mm}  
\setlength{\abovecaptionskip}{2mm}
\caption{\small An Example of errors in default ground truths of the Landsat 8 images: (a) True-color image, (b) Default ground truth for clouds(c) Icy/snowy regions, which are erroneously labeled as cloud, are highlighted with red, (d) Corrected ground truth using snow/ice removal framework.\label{Fig:just_samples}}
\end{figure}
\begin{figure*}[t]

\begin{minipage}{1\textwidth}
  \centering
  \centerline{\includegraphics[width = 175mm]{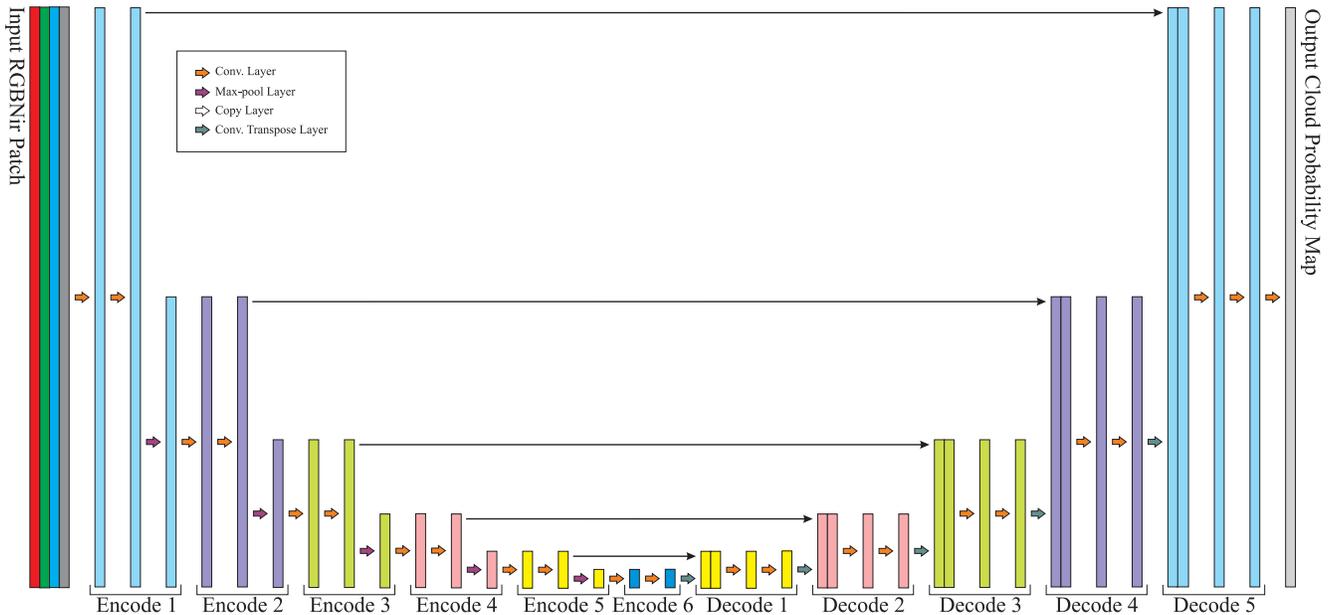}}{\vspace{2mm}}

\end{minipage}

\caption{\small The proposed network for detecting clouds. The depth of the feature map in encoding path is increased from 4 (input channels are RGBNir) to 1024. This depth is then decreased from 1024 to 1 (gray scale probability map) in the decoding path. Meanwhile, the spatial size of the feature map is reduced from $192\times 192$ to $6\times 6$ in the encoding path and, then, it is increased to $192 \times 192$ in the decoding path. The copy layer between Encode $i$ block and Decode $j$ block, concatenates the output of second convolution layer in Encode $i$ block to the output of convolution transposed layer in Decode $j$ block. 
\label{Fig:Arch}}

\end{figure*}

Function of Mask (FMask) \cite{fmask} and Automated Cloud-Cover Assessment (ACCA) \cite{acca} algorithms are among the most known and reliable threshold-based algorithms for cloud identification. They use a decision tree to label each pixel as cloud or non-cloud. In each branch of the tree, the decision is made based on the result of a thresholding function that utilizes one or more spectral bands of data. Haze optimized Transformation (HOT) is among the group of handcrafted methods, which isolates haze and thick clouds from other pixels using the relationship between spectral responses of two visible bands. \cite{bag}, as another handcrafted approach, incorporates an object-based Support Vector Machine (SVM) classifier to identify clouds from non-cloud regions using local cloud patterns. With the recent advances made in deep-learning algorithms for image segmentation, several methods have been developed for cloud detection using deep-learning approaches.  Xie et al. \cite{multilevel} trained a convolutional Neural network (CNN) from multiple small patches. This network classified each image patch into three classes of thin cloud, thick cloud, and non-cloud and as the output it created a probability map for each class. A major problem in cloud detection based on deep-learning is the lack of accurately annotated ground truth. Most default ground truths, obtained through automatic/semi-automatic approaches are not accurate enough. For instance, they label icy or snowy areas as clouds. Such erroneous ground truth limits their use for training new systems based on deep-learning. Fig. \ref{Fig:just_samples} illustrates an example of these errors in a default ground truth.

Although the above mentioned methods have shown limited good results for scenes including thick clouds, they cannot deliver robust and accurate results in scenes where snow is present alongside of the cloud.

Here, we propose a new method based on both thresholding and deep-learning to identify the cloud regions and separate them from icy/snow ones in multi-spectral Landsat 8 images. Our threshold based method utilizes band 2 in Landsat 8 and image gradient to detect regions of snow. We augment the existing Landsat 8 ground truth images by first identifying the icy/snowy regions and second removing them from the ground truth data that is used for the training of our deep-learning system.  Our proposed deep-learning system is a Fully Convolutional Neural Network (FCN) that is trained using cropped patches of the training set images. The weights of the trained network are used to detect cloud pixels in an end-to-end manner. Unlike FMask and ACCA, this approach is not blind to the existing global and local cloud contexts in the image. In addition, since only four spectral bands---Red, Green, Blue (band 2), and Nir (RGBNir)---are required for the system training and prediction, this architecture can be simply utilized for detection of clouds in images obtained from many other satellites as well as air-born systems.

\section{Proposed Method}
\subsection{Landsat 8 Images}
Landsat 8 multi-spectral data consists of nine spectral bands collected from Operational Land Imager (OLI) sensor and two thermal bands obtained by Thermal Infrared Sensor (TIRS) sensor each measuring a different range of wavelengths. Table I summarizes the specification of these bands. In this paper, we only use four spectral bands—Band 2 to Band 5. Also, there is a Quality Assessment (QA) band, which is developed by the Landsat 8 Cloud Cover Assessment (CCA) system and the FMask algorithm\cite{landsat_handbook}. The default cloud/snow ground truths of an image can be extracted from QA band.


\renewcommand{\arraystretch}{1.1}
\begin{table}
\footnotesize
\begin{minipage}[t]{0.5\textwidth}
\centering
\caption{Landsat 8 Spectral Bands.\label{Tab:landsat8_bands}} 
\begin{tabular}{|c|c|}
\hline
\textbf{Spectral Bands}               & \textbf{Wavelength (um)} \\ \hline
Band 1 - Ultra Blue & 0.435 - 0.451                                  \\ \hline
Band 2 - Blue                         & 0.452 - 0.512           \\ \hline
Band 3 - Green                        & 0.533 - 0.590            \\ \hline
Band 4 - Red                          & 0.636 - 0.673             \\ \hline
Band 5 - Near Infrared (Nir)          & 0.851 - 0.879             \\ \hline
Band 6 - Shortwave Infrared 1  & 1.566 - 1.651            \\ \hline
Band 7 - Shortwave Infrared 2  & 2.107 - 2.294            \\ \hline
Band 8 - Panchromatic                 & 0.503 - 0.676             \\ \hline
Band 9 - Cirrus                       & 1.363 - 1.384           \\ \hline
Band 10 - Thermal Infrared (TIRS) 1   & 10.60 - 11.19            \\ \hline
Band 11 - Thermal Infrared (TIRS) 2   & 11.50 - 12.51            \\ \hline
\end{tabular}
\end{minipage}
\end{table}


\subsection{Snow/Ice Removal Framework}
To augment/correct the cloud ground truths of the Landsat 8 training data we, first, apply a snow/ice removal approach. To do so, each Landsat 8 spectral band image is first divided into three distinct regions; snow, cloud, and clear using the information provided with Landsat 8 QA band. The gradient magnitude for each pixel, is then obtained. Once calculated, the average image gradient magnitude for each of the snow, cloud, and clear regions is determined. Comparing these averages in four spectral bands reveals a considerable difference between the snow region and the rest of the image. Since Band 2 exhibited the greatest proportional difference between the average gradient magnitude of the snow region and rest of the image, we utilized this band for snow/ice removal framework. After determining the image gradient of Band 2, a global threshold is applied to isolate pixels with greater values and produce a binary snow mask. By removing detected snow regions from the default cloud ground truth extracted from Landsat 8 QA band, a corrected and more accurate binary cloud mask is obtained. Fig. \ref{Fig:just_samples}(d) illustrates a corrected cloud ground truth image using the above snow/ice removal framework.

\subsection{Cloud Detection Framework}
Once the ground truths are corrected, we utilize them in a deep-learning framework to identify cloud pixels in an image. In FCNs the spatial size of the output image is same as the input image. This characteristic allows these type of CNNs to be used in pixel-wise labeling tasks such as image segmentation. The proposed CNN in this paper has a FCN architecture, which is inspired by U-Net \cite{UNet}. U-Net is introduced to segment specific regions in Electron Microscopic (EM) stack images. This network is widely used in many other computer vision applications \cite{3DUNet,finger_prnt_UNet}. It basically has a fully convolutional encoder (contracting) path, which is connected to a fully convolutional decoder (expanding) path. Some skip connections attaches the encoding blocks in contracting path to the analogous decoding blocks in the expanding path.

The block diagram of the proposed network is shown in Fig. \ref{Fig:Arch}. It has six encoding and five decoding blocks. In each of these blocks there are two convolution layers to extract the semantic features of the image. In a convolution layer, a $3\times3$ kernel is convolved with the input of the layer. A Rectified Linear Unit (ReLU) \cite{ReLU} is then applied to generate the output. In the encoding path, the output of a convolution layer is followed by a maxpooling layer to reduce the spatial size of the feature map. In the decoding path, the spatial size of the feature map is gradually increased to reach to the original input size of the network using convolution transposed layers in decoding blocks. Image features, obtained from an encoding block, is utilized in the analogous decoding block---using a copy layer. By applying repetitive encoding and decoding blocks, low-level features of the image at the very beginning layers of the network are evolved to high-level semantic contexts at the output probability map of the network.     

The spatial dimensions of the input images in the proposed network is $192\times192\times4$ pixels. Since each of the spectral band of the Landsat 8 is very large---in order of $8000\times8000$ pixels---we have to cut them into smaller image patches. Therefore, each spectral band image is cropped into $384\times384$ non-overlapping patches. Before training, these patches are resized to $192\times192$. Then four patches corresponding to Red, Green, Blue, and Nir bands are stacked on the top of each other to create a 4D input and then this input is fed to the network. To reduce the vulnerability of the approach to misleading patterns with similarities to the cloud patterns, we augmented the input patches with geometric transformations such as horizontal flipping, rotation, and zooming. In the very last convolution layer of the network, a $sigmoid$ activation function is utilized to extract the output probability map. The following soft \textit{Jaccard} loss function \cite{jacc1,jacc2} is implemented to optimize the network through Adam gradient descent \cite{ADAM} approach:

\begin{equation}
\large
\begin{split}
L(h,y) \! = \!-\dfrac{\sum\limits_{i=1}^{n} h_{i} y_i+\epsilon}{\sum\limits_{i=1}^{n} h_{i} + \sum\limits_{i=1}^{n} y_i - \sum\limits_{i=1}^{n} h_{i} y_i+\epsilon},
\\ 
\end{split}
\label{Eq:loss}
\end{equation}
Here, $h$ is the ground truth and $y$ is the probability map that is obtained from output of the \textit{sigmoid} function in the last layer of the network. $n$ is the total number of pixels in the ground truth. $y_i$ and $h_{i}$ are the $i$th pixel of $y$ and $h$. $\epsilon$ is a small real number to avoid division by zero.
The learning process is started from the weights that are constructed from a uniform random distribution between $[-1, 1]$. We set the initial learning rate of the training as $10^{-4}$. 

The training process is done for 600 epochs. After this number of epochs the network is converged to an appropriate local minimum. The obtained weights are then utilized for the prediction purposes. Before prediction, non-overlapping $384\times384$ patches are extracted from each of the four spectral bands of the given test image. Then these patches are resized to $192\times192$ and stacked together. Once the cloud features corresponding to each patch are obtained, the output cloud probability map is resized to $384\times384$ pixels. These resized patches are then stitched up together to create a cloud probability map for the entire image. By doing a simple thresholding the binary cloud mask of the input image is obtained.
\begin{figure*}[t]
\centering

\begin{minipage}{0.23\textwidth}
\centering
\centerline{\includegraphics[height=43mm, width=43mm]{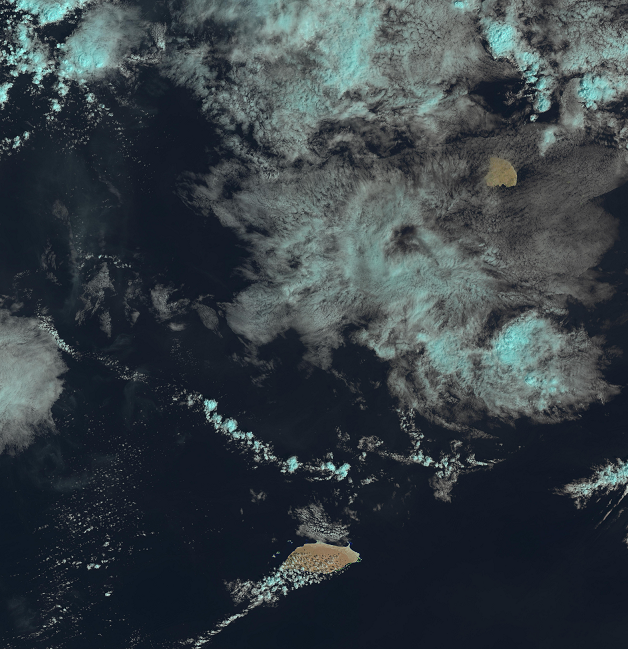}}\footnotesize{(a)}

\end{minipage}
\vspace{1mm}
\hspace{1mm}
\begin{minipage}{0.23\textwidth}
  \centering
  \centerline{\includegraphics[height=43mm, width=43mm]{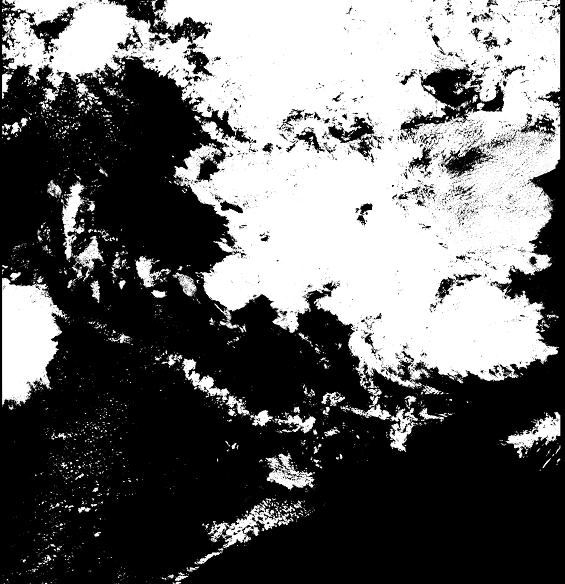}}\footnotesize{(b)}

\end{minipage}
\vspace{1mm}
\hspace{1mm}
\begin{minipage}{0.23\textwidth}
  \centering
  \centerline{\includegraphics[height=43mm, width=43mm]{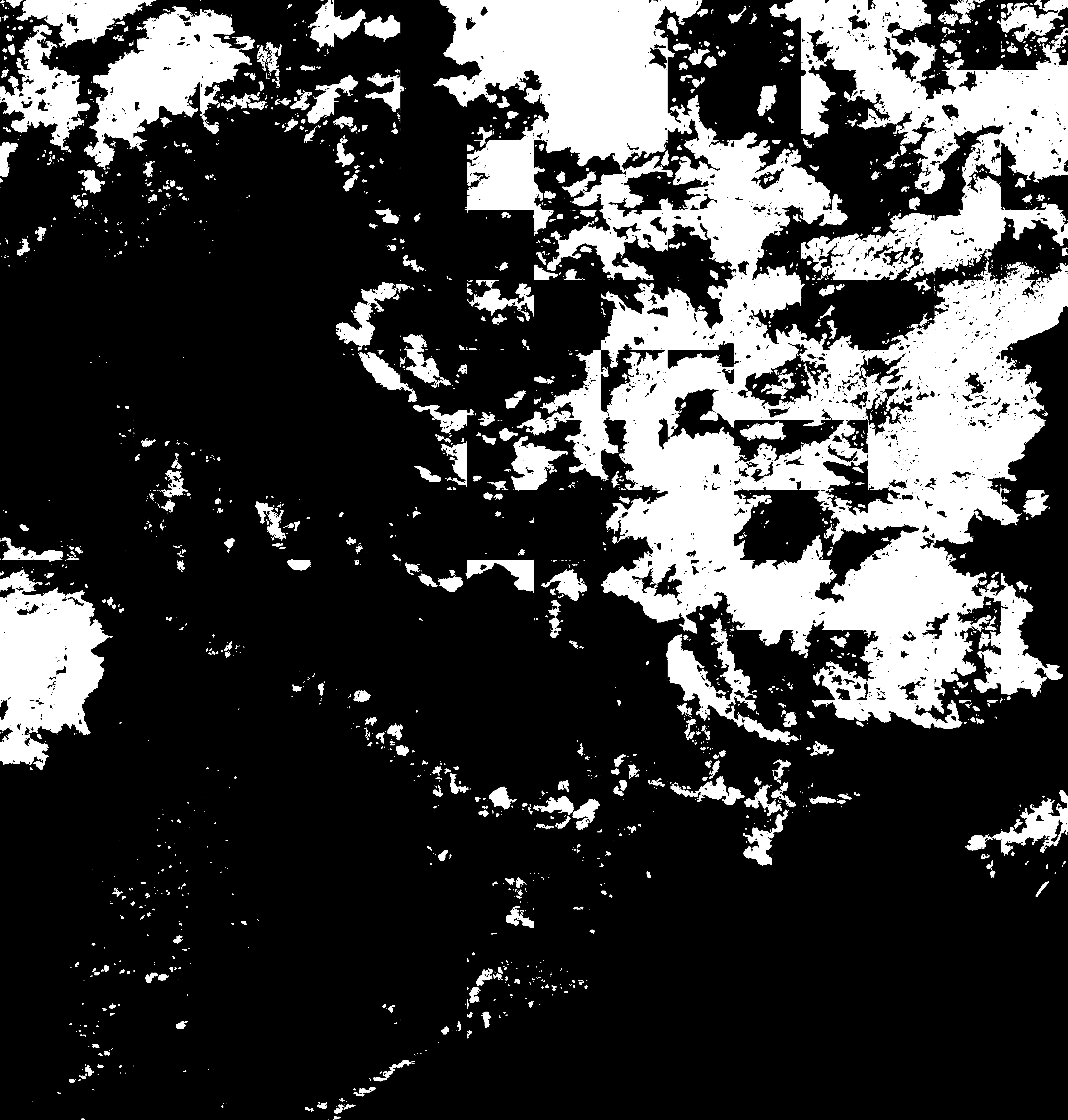}}\footnotesize{(c)}

\end{minipage}
\vspace{1mm}
\hspace{1mm}
\begin{minipage}{0.23\textwidth}
\centering
\centerline{\includegraphics[height=43mm, width=43mm]{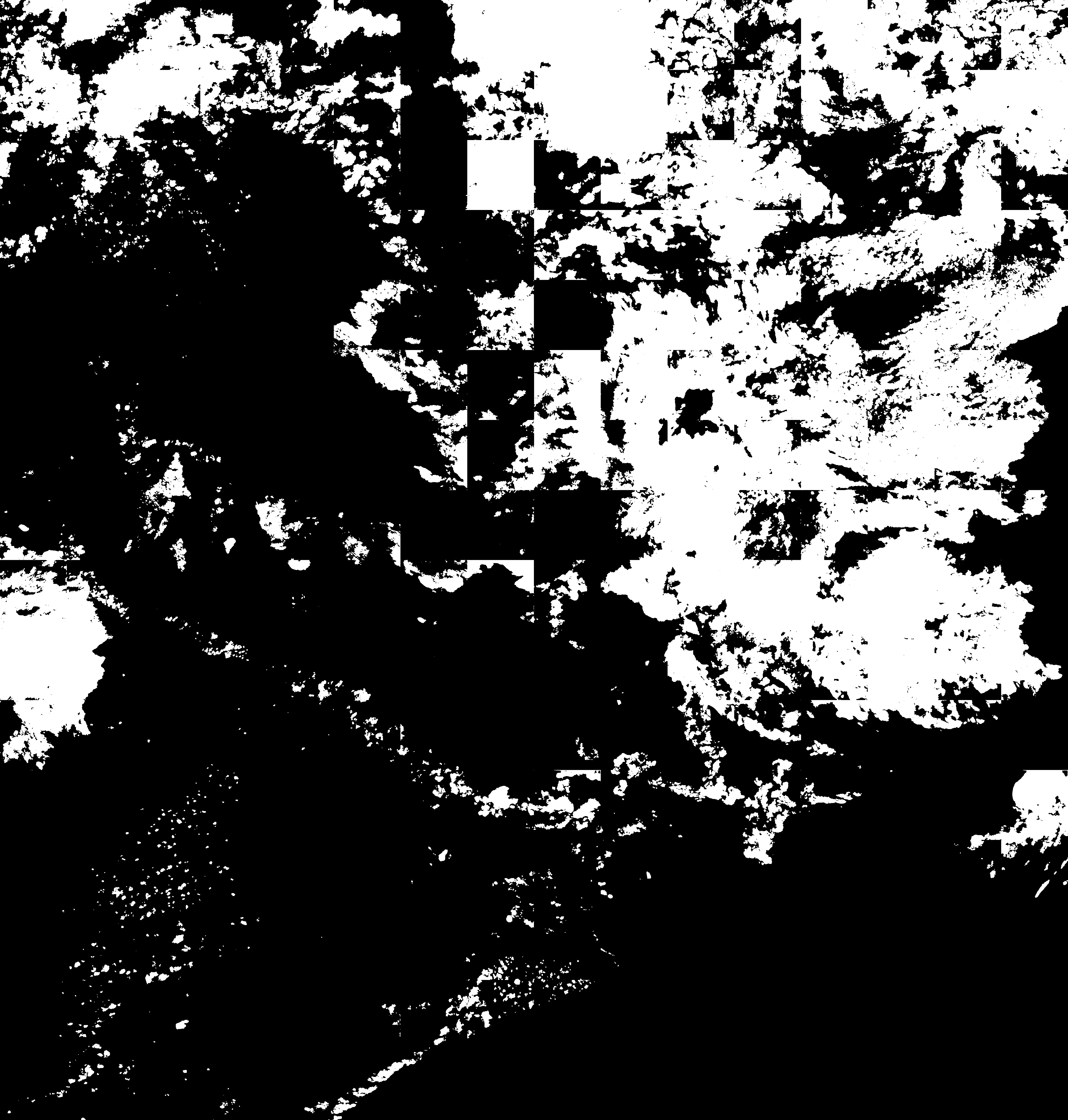}}\footnotesize{(d)}

\end{minipage}


\begin{minipage}{0.23\textwidth}
\centering
\centerline{\includegraphics[height=43mm, width=43mm]{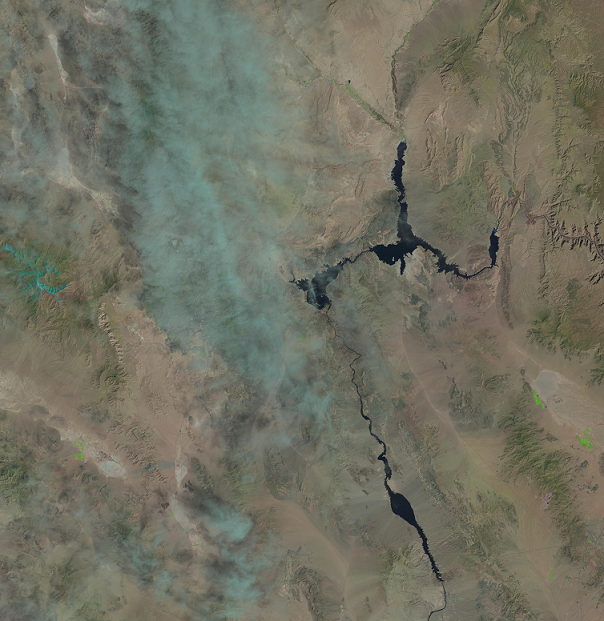}}\footnotesize{(e)}

\end{minipage}
\vspace{1mm}
\hspace{1mm}
\begin{minipage}{0.23\textwidth}
  \centering
  \centerline{\includegraphics[height=43mm, width=43mm]{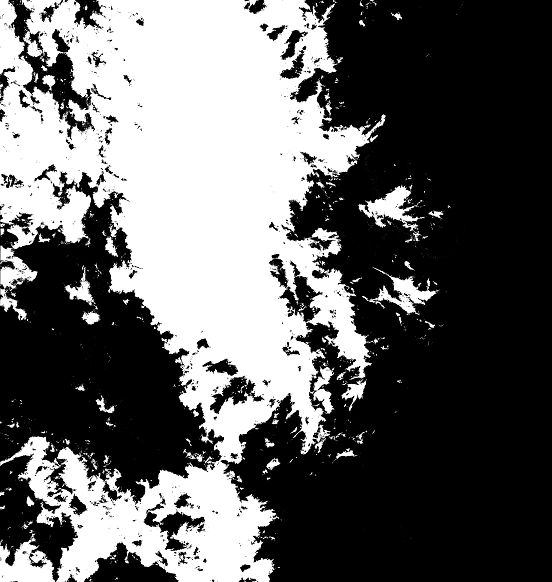}}\footnotesize{(f)}

\end{minipage}
\vspace{1mm}
\hspace{1mm}
\begin{minipage}{0.23\textwidth}
  \centering
  \centerline{\includegraphics[height=43mm, width=43mm]{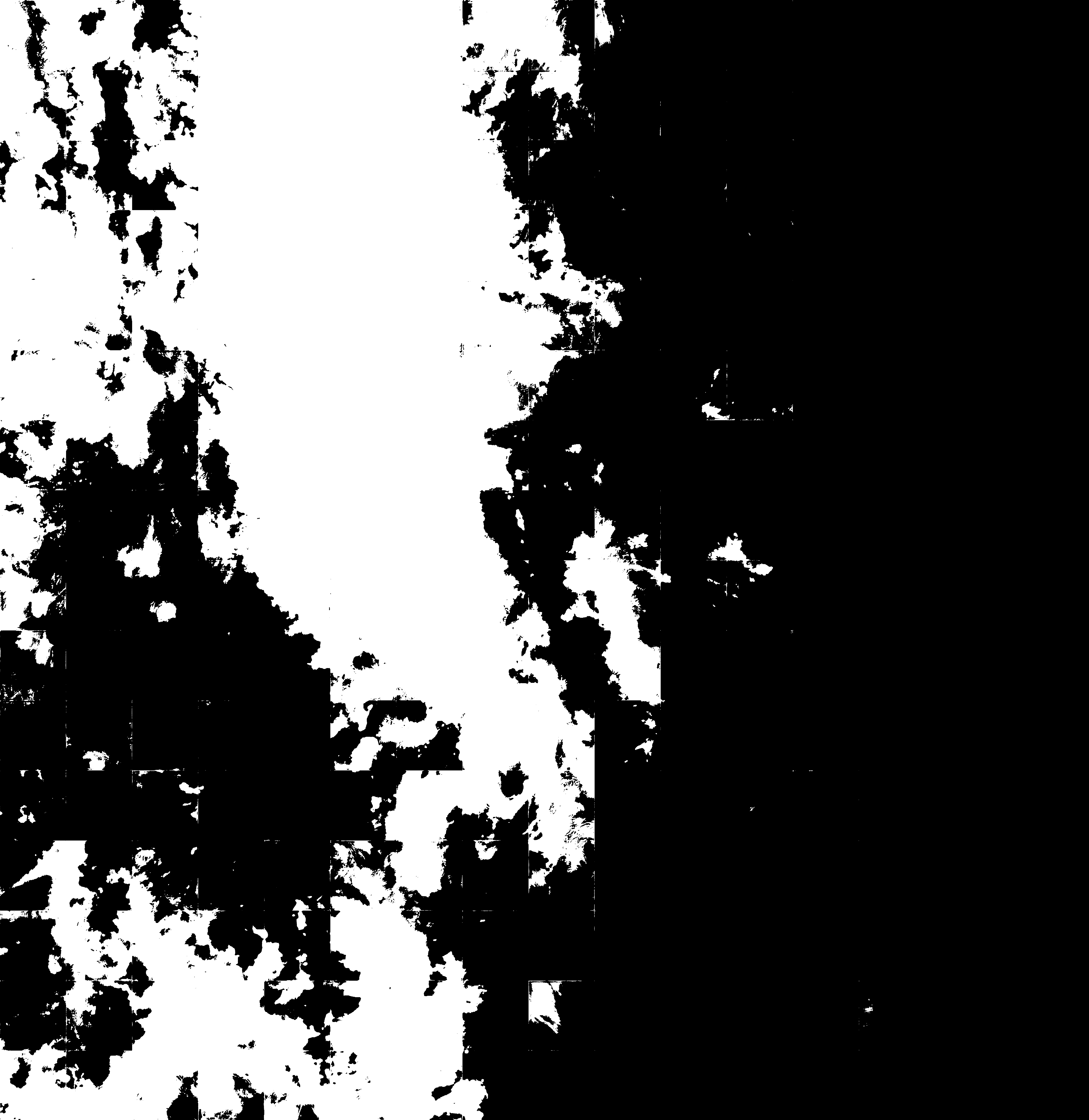}}\footnotesize{(g)}

\end{minipage}
\vspace{1mm}
\hspace{1mm}
\begin{minipage}{0.23\textwidth}
\centering
\centerline{\includegraphics[height=43mm, width=43mm]{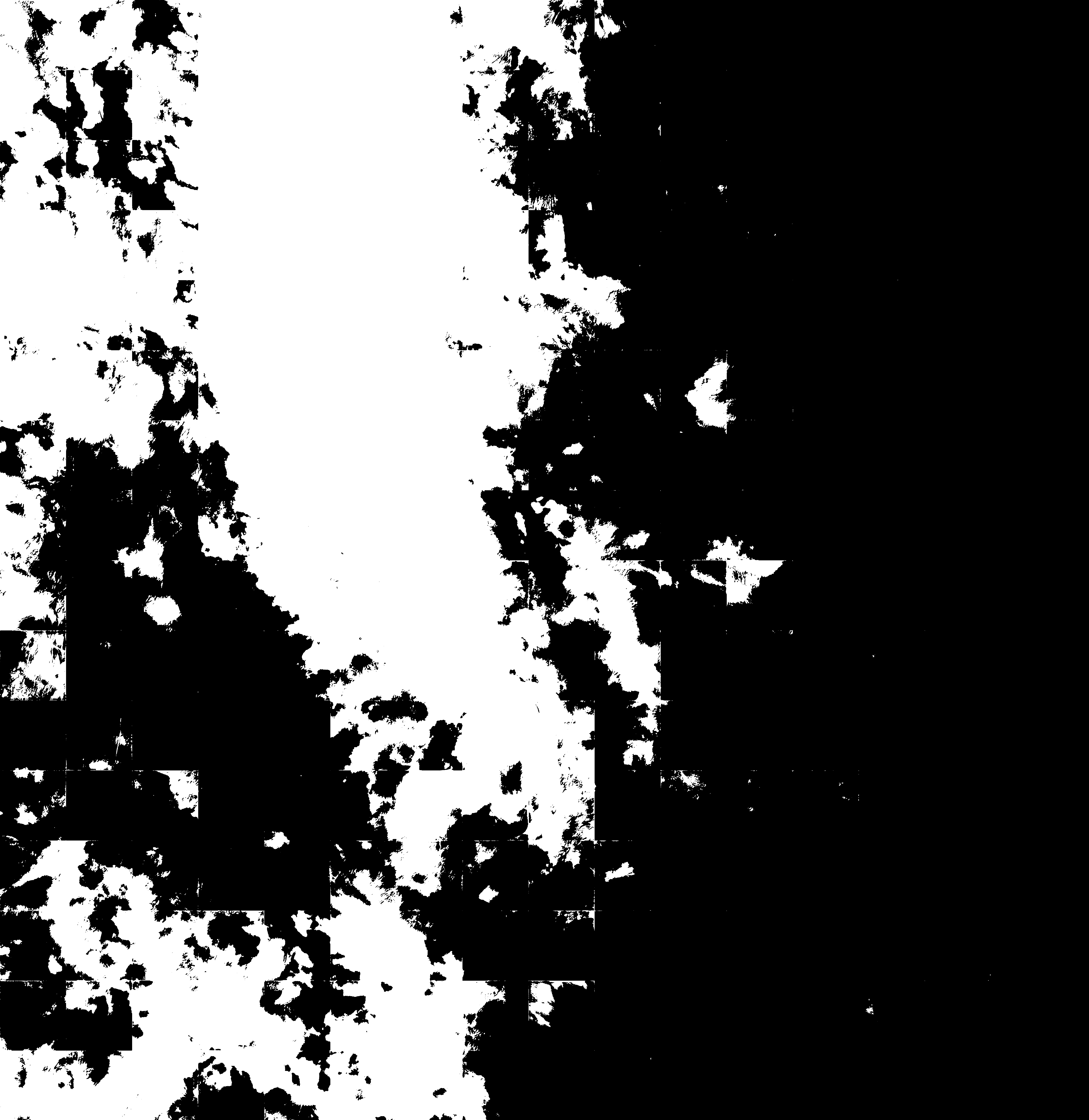}}\footnotesize{(h)}

\end{minipage}

\vspace{-1mm}
\hspace{-2.5mm}  
\setlength{\abovecaptionskip}{2mm}
\caption{\small Examples of the cloud masks obtained by the proposed method: (a),(e) True-color input images, (b), (f) Manual ground truths, (c), (g) Predicted cloud mask without snow/ice correction, (d), (h) Predicted cloud masks with snow/ice correction.
\label{Fig:experimental_samples}}
\end{figure*}


\renewcommand{\arraystretch}{1.1}
\begin{table*}
\small
\begin{minipage}[t]{1\textwidth}
\centering
\caption{System performance measures (in~\%).\label{Tab:numerical}} 

\begin{tabular}{|c|c|c|c|c|}
\hline
\textbf{Method}              & \textbf{Jaccard}                       & \textbf{Precision}   & \textbf{Recall} & \textbf{Overall Accuracy}   \\ 
\hhline{|=|=|=|=|=|}
FCN without snow/ice correction  &          62.63    & 72.59 &  79.39&        87.81 \\ \hline
FCN with snow/ice correction                           & \textbf{65.36} &  \textbf{73.54}&  \textbf{82.26}  & \textbf{88.30}  \\ \hline
Improvement Percentage                        &   4.36 & 1.30 & 3.62     & 0.56 \\ \hline

\end{tabular}
\end{minipage}
\end{table*}

\section{Experimental Results}
\subsection{Dataset}
We have created a new dataset for cloud detection purposes. This dataset includes two sets of training and test images. The training set contains 4600 patches that are cut from 18 Landsat 8 images. Each image has 4 bands. It also includes ground truth patches (extracted from the Landsat 8 QA band) which we refer to it by the default ground truth. In the test side, the test set holds 5100 patches (obtained from 20 images that are different from those for the training set) with 4 bands. Along with these image patches, we have the manually created ground truth of the clouds. 

Before training our system, we have applied our proposed threshold-based snow/ice removal method (Section \Rmnum{2}.B) to automatically identify the snowy/icy regions and remove them from the default ground truths. We then trained our system twice once using this corrected ground truths and once using the default ground truths. In both cases we run the system on the test set and compare the outputs with the manually created ground truths to highlight the improvement. It is important to mention that both the training and the test images are selected to cover many scene elements such as vegetation, bare soil, buildings, urban areas, water, snow, ice, haze, different types of cloud patterns, etc. and the average percentage of cloud coverage in both of the train and test sets is around 50\%. This dataset is publicly available to the community by request.

\subsection{Evaluation Metrics}
The performance of the proposed algorithm is determined by evaluating the overall accuracy, recall, precision, and \textit{Jaccard} index for the masks it produces. These measures are defined as follows:
 
\begin{equation}
\small
\begin{split}
&\quad \text{Jaccard Index}=\frac{TP}{TP+FN+FP}, \\
&\quad \text{Precision}=\frac{TP}{TP+FP}, \\
&\quad \text{Recall}=\frac{TP}{TP+FN}, \\
&\quad \text{Overall Accuracy}=\frac{TP+TN}{TP+TN+FP+FN}, \\
\end{split}
\vspace{5mm}
\label{Eq:Ev}
\end{equation}
Where TP, TN, FP, and FN are true positive, true negative, false positive, and false negative respectfully. The \textit{Jaccard} index relates both recall and precision and is a measure of the similarity between two sets.

\subsection{Numerical and Visual Results}
Table \ref{Tab:numerical} demonstrates experimental results of the proposed method on our test set. As shown, the \textit{Jaccard} index of the cloud masks obtained by augmented ground truths is improved by 4.36\%. This indeed highlights the effectiveness of the proposed snow/ice removal framework in our proposed method. Also recall measure is increased by 3.62\%. This measure indicates that the number of cloud pixels labeled correctly as cloud is increased. Some visual examples of the predicted shadow masks from sample images of  our test set are displayed in Fig.\ref{Fig:experimental_samples}.

\section{Conclusion}
In this paper a deep-learning based approach is proposed to detect the cloud pixels in Landsat 8 images using only four spectral bands of RGBNir. Our pixel-level segmentation framework extracts the semantic local and global features of the clouds in an image by a high accuracy. This framework can be utilized for other segmentation tasks in the applications of remote sensing images of satellites or airborne sensors. We also introduce a novel cloud detection dataset with accurately annotated cloud pixels. In our future work, we will focus on improving the network’s field of view to identify more cloud context out of the images.



\balance


\bibliographystyle{IEEEtran}
\bibliography{refs}

\end{document}